# Off-the-Shelf Neural Network Architectures for Forex Time Series Prediction come at a Cost

Theodoros Zafeiriou

School of Science & Technology, Hellenic Open University, Patras, Greece, zafiriou.theodoros@ac.eap.gr

Dimitris Kalles

School of Science & Technology, Hellenic Open University, Patras, Greece, kalles@eap.gr

## ABSTRACT

Our study focuses on comparing the performance and resource requirements between different Long Short-Term Memory (LSTM) neural network architectures and an ANN specialized architecture for forex market prediction. We analyze the execution time of the models as well as the resources consumed, such as memory and computational power. Our aim is to demonstrate that the specialized architecture not only achieves better results in forex market prediction but also executes using fewer resources and in a shorter time frame compared to LSTM architectures. This comparative analysis will provide significant insights into the suitability of these two types of architectures for time series prediction in the forex market environment.

## 1 Introduction

In recent years, the application of artificial neural networks (ANNs) in financial forecasting, particularly in the realm of forex (foreign exchange) market prediction, has garnered significant attention, due to their ability to capture complex patterns in time series data. Among the various types of ANNs, Long Short-Term Memory (LSTM) networks have emerged as a popular choice for modeling sequential data and exhibiting strong predictive capabilities.

However, while LSTM networks have demonstrated promising results in forex market prediction, there remains a need to evaluate their performance in terms of execution time and resource utilization. Additionally, the emergence of specialized neural network architectures tailored specifically for financial time series forecasting presents an intriguing opportunity for comparison.

This study aims to provide a comprehensive comparative analysis of the execution time and resource requirements between LSTM neural network architectures and an ANN specialized architecture designed for forex market prediction. By examining both the performance and resource efficiency of these architectures, this research seeks to offer valuable insights into their suitability for time series prediction in the dynamic and volatile environment of the forex market.

## 2 A brief background on the comparison of artificial neural network architectures in time series prediction with respect to resource consumption

David Salinas et al [1] present DeepAR, a carefully designed model for time series forecasting based on autoregressive recurrent networks. DeepAR is used for probabilistic forecasting, providing predictions accompanied by uncertainty estimates. The model is scalable to multiple time series and can incorporate external variables. The paper provides a comparison of DeepAR with other time series forecasting methods such as traditional autoregressive models (e.g., ARIMA). It mentions the computational resources required for training and prediction with DeepAR compared to other approaches.

Bryan Lim et al [2] introduce Temporal Fusion Transformers (TFT), a novel model for time series forecasting based on the attention-based Transformer architecture. TFTs enable interpretable predictions and can forecast multiple horizons simultaneously. The paper provides a comparison of the computational resources required by TFTs compared to other time series forecasting models such as RNNs and ARIMA. It discusses the effectiveness of TFTs in relation to the resources required for training and prediction.

J. Sevilla et al [3] investigate the growth of computational requirements for training ML models. Before 2010, training compute doubled roughly every 20 months, following Moore's law. Since the rise of Deep Learning in the early 2010s, compute scaling accelerated, doubling approximately every 6 months. A new trend emerged in late 2015 with large-scale ML models requiring 10 to 100 times more compute. The study splits ML compute history into three eras: Pre-Deep Learning, Deep Learning, and Large-Scale Era

Another work [4] investigates the extent of deep learning's dependency on computing power. It reveals that progress in various applications heavily relies on increased computational resources. However, this trajectory is becoming unsustainable. Future progress will require more computationally-efficient methods or alternative machine learning approaches.

Daniel Justus et al [5] discuss the challenge of accurately predicting the training time for deep learning networks. Despite deep learning's superior performance, estimating the time required to train a network remains elusive. Training time depends on both the per-epoch duration and the total number of epochs needed to achieve desired accuracy. Existing methods often assume a linear relationship between training time and floating-point operations, but this assumption breaks down when other factors dominate execution time (e.g., data loading or suboptimal parallel execution). The proposed alternative approach trains a deep learning network to predict execution times for individual parts, which are then combined to estimate overall execution time. This method models complex scenarios and can predict execution times for unseen scenarios, aiding hardware and model selection.

Another work [6] addresses the challenge of estimating the number of optimization steps required for a pre-trained deep network to converge to a specific loss value. Leveraging the fact that fine-tuning dynamics resemble those of a linearized model, the authors approximate training loss and accuracy using a low-dimensional Stochastic Differential Equation (SDE) in function space. This allows them to predict the time it takes for Stochastic Gradient Descent (SGD) to fine-tune a model without actual training. Their method achieves a 20% error margin for ResNet training time across various datasets and hyperparameters, at a significantly reduced computational cost compared to real training.

Finel, F. et al [7] present a procedure for designing a DNN that estimates execution time for training deep neural networks per batch on GPU accelerators. The estimator is intended for shared GPU infrastructures, providing estimated training times for various network architectures when users submit training jobs. A co-evolutionary approach is used to fit the estimator, evolving the training set for better accuracy.

Another work [8] proposes a novel CNN architecture for classifying time series data. Instead of a single output, it introduces intermediate outputs from different hidden layers to control weight adjustments during training. These intermediate targets improve method performance, achieving higher accuracy compared to the base CNN method. The proposed CNN-TS also outperforms classical machine-learning methods and is significantly faster in training time than ResNet

Simone Bianco et al [9] provide an in-depth analysis of the majority of deep neural networks (DNNs) proposed for image recognition. It observes multiple performance indices for each DNN, such as recognition accuracy, model complexity, computational complexity, memory usage, and inference time. The study is conducted on two different computer architectures, allowing a direct comparison between DNNs running on machines with very different computational capacit

Another review paper provides an overview of long short-term memory (LSTM) networks for time series forecasting. It discusses the architecture and training process of LSTM networks, as well as their applications in various forecasting tasks. The paper analyzes the computational resources required for training and inference with LSTM networks for time series forecasting and compares them to other recurrent neural network architectures, highlighting the advantages of LSTMs in capturing long-term dependencies.

G. Wei et al comprehensive study investigates various neural network architectures for time series forecasting. It explores feedforward neural networks, recurrent neural networks, and convolutional neural networks, comparing their performance on different time series datasets. The paper provides insights into the computational resources required by different neural network architectures for time series forecasting tasks and compares their efficiency and accuracy in capturing temporal dependencies.

Laith Alzubaidi er al [10] provide a comprehensive survey of deep learning concepts, focusing on convolutional neural networks (CNNs). It outlines the importance of deep learning, describes various deep learning techniques and networks, and presents the development of CNNs architectures. the paper also discusses challenges, suggested solutions, major deep learning applications, and computational tools. It serves as a holistic starting point for understanding deep learning and its recent enhancements

These works offer a broad analysis of the performance and resource consumption of various neural network models for time series prediction and can serve as a valuable source of information for further research in this field.

In our previews work [11] we designed and implemented a series of LSTM neural network architectures which are taken as input the exchange rate values and generate the short-term market trend forecasting signal and an ANN specialize architecture based on technical analysis indicator simulators. We performed a comparative analysis of the results and came to useful conclusions regarding the suitability of each architecture and the cost in terms of time and computational power to implement them. The ANN custom architecture produces better prediction quality with higher sensitivity using fewer resources and spending less time than LSTM architectures.

The ANN custom architecture appears to be ideal for use in low-power computing systems and for use cases that need fast decisions with the least possible computational cost. The aim of this work is the comparative analysis of the results in terms of execution time and production of the prediction for the different architectures of artificial neural networks in the same environment of computing resources

## 3   Description of Prediction Models and of the Experimental Environment

In this section, we provide a brief description of the LSTM neural network architectures and the ANN custom architecture based on technical analysis indicator simulators from our previous work [11]. Additionally, we outline the data, resources, and experimentation environment for each of these.

## 3.1 Computational Resources for Experimentation

To ensure the comparability of results, we utilized the same local resources of a laptop computer for our experimentation with all neural networks:

- **Processor**: AMD Ryzen 5 5500U with Radeon Graphics @ 2.10 GHz
- **Installed RAM:** 16.0 GB (13.8 GB usable)
- **System Type:** 64-bit operating system, x64-based processor
- Operating System Version: Windows 11 Home, 23H2

The LSTM networks were executed directly in the Python 3.11 environment [12] to ensure the best performance for the aforementioned hardware resources

The ANN custom architecture[11] based on technical analysis indicator simulators was executed as a JAR file [13] using local resources.

This setup ensures consistency and comparability of results across various experimental procedures applied to different neural network architectures.

## 3.2 Prediction Models for Experimentation

We selected eight different LSTM architectures for our experimentation with parameters as shown in Table 1. All these LSTM architectures follow the sequential model and have a ReLU activation function. For more details, please refer to our previous work [11].

| Name | LSTM Units | Dense Units | Lookback * | Bidirectional | Convolutional |
| --- | --- | --- | --- | --- | --- |
| sLSTM-1-1 | 100 | 1 X 1 | 1 | No | No |
| sLSTM-15-1 | 100 | 1 X 1 | 15 | No | No |
| sLSTM-15-1,15 | 100 | 1 X 15, 1 X 1 | 1 | No | No |
| biLSTM-1-1 | 100 | 1 X 1 | 1 | Yes | No |
| biLSTM-15-1 | 100 | 1 X 1 | 15 | Yes | No |
| biLSTM-15-1,15 | 100 | 1 X 15, 1 X 1 | 15 | Yes | No |
| convLSTM-1-1 | 60 | 1 X 1 | 1 | No | Yes |
| convLSTM-1-1,15 | 64 | 1 X 1 | 15 | No | Yes |

* The number of sequences of input LSTM will train before generating an output.

**Table 1: Selected LSTM architectures**

The above methods will be compared in terms of the time they consume on the same computational resources as the **custom ANN architecture based on technical analysis indicator simulators**, as described in our previous work. Here's a concise summary of the ANN [11] architecture based on technical analysis indicator simulators:

1. **Objective:** The goal is to create an efficient prediction system for short-term trading based on technical indicators.
2. **Technical Indicators** [14] Modified arithmetic moving averages (MAs) over different price intervals, RSI-300 oscillator, CCI-300 oscillator, Williams-300 oscillator, and Price Oscillator (MA-300, MA-600, MA-900) are used.
3. **Input Parameters:** Exchange rates, time, and dates are considered.
4. Simulation Process:
    - Custom technical indicator simulators generate outputs based on input data.
    - These outputs feed into the input neurons of an Artificial Neural Network (ANN) system.
    - The ANN system consists of two sets of ANNs operating in pairs.
    - One ANN (back-propagation mode) aligns with trend prediction using past values.
    - Its learned weights transfer to its paired ANN (feed-forward mode), which predicts current data.
    - All feed-forward ANNs combine to generate the final trend forecast.
5. **Architecture Modification:** Inspired by Generative Adversarial Networks (GANs) [15], this architecture enhances prediction accuracy.

This ANN-based system aims to optimize ultra-short-term trading decisions by simulating human expert judgment and adapting to changing market conditions. The technical indicators and neural network structure play crucial roles in achieving accurate predictions.

## 3.3 Selection of the Exchange Rate and Experimental Data Source

In our research, we deliberately focused on the **EUR/USD exchange rate**, which holds a prominent position as the world's most significant trading currency pair. The substantial market depth of this pair serves as a safeguard against any attempts by interest groups to manipulate prices and distort the true representation.

To gather experimental data, we turned to TrueFx [16] one of the industry's leading forex data servers.

The dataset we analyzed covers the tick-to-tick EUR/USD exchange rate for the months of **October, November, and December 2021**. Initially, this dataset contains over **10 million values**, which we meticulously pre-processed to remove flat areas where the exchange rate remained constant.

# 4 Experimentation and Results

## 4.1 Presentation and Analysis of Results

Below, we provide a brief overview of the performance of our experimentation architectures. Detailed experimentation results appear in [11].

|  | OCTOBER | | NOVEMBER | | DECEMBER | |
|---|---|---|---|---|---|---|
| **ANN** | STA | STS | STA | STS | STA | STS |
| Successful Forecasting Signals | 3808 | 310 | 10923 | 880 | 10989 | 437 |
| Total forecasting signals | 4641 | 407 | 13371 | 1070 | 13689 | 593 |
| % Success | **82,05%** | **76,17%** | **81,69%** | **82,24%** | **80,28%** | **73,69%** |
| **sLSTM-1-1** | | | | | | |
| Successful Forecasting Signals | 761 | 101 | 831 | 161 | 1419 | 253 |
| Total forecasting signals | 1091 | 161 | 1133 | 237 | 1921 | 424 |
| % Success | **69,75%** | **62,73%** | **73,35%** | **67,93%** | **73,87%** | **59,67%** |
| **sLSTM-15-1** | | | | | | |
| Successful Forecasting Signals | 769 | 96 | 483 | 80 | 1334 | 224 |
| Total forecasting signals | 1122 | 158 | 653 | 115 | 1803 | 372 |
| % Success | **68,54%** | **60,76%** | **73,97%** | **69,57%** | **73,99%** | **60,22%** |
| **sLSTM-15-1,15** | | | | | | |
| Successful Forecasting Signals | 782 | 100 | 310 | 58 | 1393 | 248 |
| Total forecasting signals | 1133 | 164 | 416 | 80 | 1892 | 418 |
| % Success | **69,02%** | **60,98%** | **74,52%** | **72,50%** | **73,63%** | **59,33%** |
| **biLSTM-1-1** | | | | | | |
| Successful Forecasting Signals | 779 | 105 | 760 | 142 | 1413 | 249 |
| Total forecasting signals | 1122 | 167 | 1033 | 213 | 1915 | 420 |
| % Success | **69,43%** | **62,87%** | **73,57%** | **66,67%** | **73,79%** | **59,29%** |
| **biLSTM-15-1** | | | | | | |
| Successful Forecasting Signals | 848 | 113 | 462 | 77 | 1344 | 238 |
| Total forecasting signals | 1244 | 197 | 621 | 109 | 1823 | 401 |
| % Success | **68,17%** | **57,36%** | **74,40%** | **70,64%** | **73,72%** | **59,35%** |
| **biLSTM-15-1,15** | | | | | | |
| Successful Forecasting Signals | 821 | 110 | 289 | 50 | 1397 | 259 |
| Total forecasting signals | 1199 | 191 | 378 | 68 | 1909 | 439 |
| % Success | **68,47%** | **57,59%** | **76,46%** | **73,53%** | **73,18%** | **59,00%** |
| **convLSTM-1-1** | | | | | | |
| Successful Forecasting Signals | 781 | 107 | 968 | 203 | 1350 | 240 |
| Total forecasting signals | 1125 | 169 | 1330 | 314 | 1829 | 402 |
| % Success | **69,42%** | **63,31%** | **72,78%** | **64,65%** | **73,81%** | **59,70%** |
| **convLSTM-1-1,15** | | | | | | |
| Successful Forecasting Signals | 352 | 37 | 106 | 24 | 894 | 104 |
| Total forecasting signals | 471 | 51 | 148 | 36 | 1179 | 165 |
| % Success | **74,73%** | **72,55%** | **71,62%** | **66,67%** | **75,83%** | **63,03%** |

**Table 2:** Aggregated Results of the Forecasting Quality

The table 2 presents the aggregate results for different LSTM architectures and a custom ANN architecture over three months. The results are divided into two indices: STA and STS. Success is measured by the trend accuracy of all forecast signals (STA) and, specifically, the trend accuracy of robust forecast signals (STS). A robust forecast signal is defined as one with an intensity of less than or equal to -1 or greater than or equal to 1, in accordance with the criteria set out in Table 3. A forecast signal is deemed successful if it aligns with both the direction and magnitude and is verified within a span of 900 foreign exchange rate data points, which equates to roughly 15 minutes. Below, we present a summary of the results as shown in the table:

**October**
- The ANN architecture produced a total of 4,641 forecasting signals in the STA index, with 3,808 being successful, resulting in an 82.05% success rate. In the STS index, it had a 76.17% success rate with 310 out of 407 signals being successful.
- The sLSTM-1-1 architecture had lower success rates, with the highest being 69.75% in the STA index.

**November**
- The ANN architecture's success rate increased slightly in the STA index to 81.69% and in the STS index to 82.24%.
- The sLSTM-15-1,15 architecture showed improvement, with a success rate of 74.52% in the STA index.

**December**
- The ANN architecture maintained a high success rate of over 80% in both indices.
- The convLSTM-1-1,15 architecture had the highest success rate among LSTM architectures at 75.83% in the STA index.

**Overall Performance**
- The ANN architecture consistently outperformed all LSTM architectures in both success rates and the absolute number of forecasting signals.
- By the end of the experiment, the ANN architecture produced 31,701 forecasting signals with an 81.13% success rate.
- The best-performing LSTM architecture was sLSTM-1-1, with a 72.64% success rate and 4,145 forecasting signals.
- The custom ANN architecture not only had higher success rates but also generated significantly more forecasting signals, 7.6 times more than the LSTM architectures, and 8.5 percent more successful signals.

The above summary highlights the ANN architecture's superior performance in forecasting accuracy and sensitivity compared to the LSTM architectures throughout the experiment. The data suggests that the ANN architecture is a more robust model for generating forecasting signals.

We now move on to the results of our main experimentation for this work. In Table 3 and Figure 1 we present the time in seconds required by the various compared architectures to produce the forecasting results

| **Name** | **Time in Seconds** | | | |
|---|---|---|---|---|
| | **Month** | | | **Overall** |
| | OCT 21 | NOV 21 | DEC 21 | |
| **sLSTM-1-1** | 117 | 89 | 116 | 322 |
| **sLSTM-15-1** | 227 | 170 | 226 | 623 |
| **sLSTM-15-1,15** | 233 | 179 | 232 | 644 |
| **biLSTM-1-1** | 142 | 109 | 141 | 392 |
| **biLSTM-15-1** | 282 | 212 | 279 | 773 |
| **biLSTM-15-1,15** | 287 | 217 | 290 | 794 |
| **convLSTM-1-1** | 195 | 149 | 193 | 537 |
| **convLSTM-1-1,15** | 302 | 233 | 302 | 837 |
| **Custom ANN architecture** | 32 | 24 | 32 | 88 |

**Table 3:** Aggregated Results of the time efficiency

The table compares the time efficiency of different LSTM variants and a custom ANN architecture across three consecutive months (October, November, and December 2021) and provides an overall time performance.

**Monthly Performance:**
- In October, the custom ANN architecture was the fastest, taking only 32 seconds to produce results. The slowest was the convLSTM-1-1,15, taking 302 seconds.
- November followed a similar pattern, with the custom ANN architecture requiring just 24 seconds and the convLSTM-1-1,15 again taking the longest at 233 seconds.
- December saw no significant change in the trend, with the custom ANN architecture remaining the quickest at 32 seconds, while the convLSTM-1-1,15 took the most time at 302 seconds.

**Overall Performance:**
- The custom ANN architecture demonstrated remarkable efficiency, with an overall time of 88 seconds across the three months. This is significantly lower than the other architectures, indicating a faster processing capability.
- The sLSTM-1-1 architecture was the next best performer with an overall time of 322 seconds.
- The biLSTM and convLSTM architectures showed a higher time requirement, with the biLSTM-15-1,15 taking the longest at 794 seconds overall.

**Conclusion:**
- The custom ANN architecture outperforms all LSTM variants in terms of time efficiency, taking less than a third of the time required by the fastest LSTM architecture.
- The data suggests that the custom ANN architecture is not only more time-efficient but also likely more cost-effective in terms of computational resources.

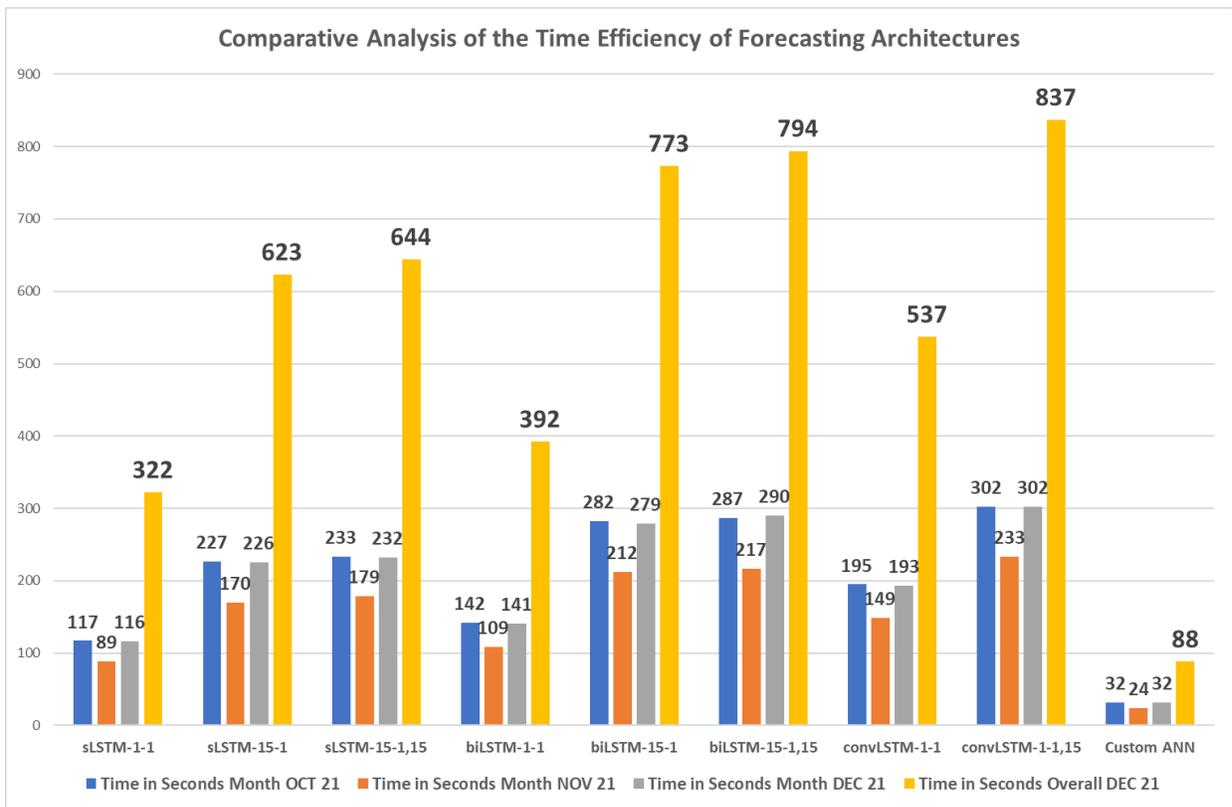

**Figure 1: Comparative Analysis of the Time Efficiency of Forecasting Architectures**

This analysis indicates that the custom ANN architecture could be a superior choice for tasks where time efficiency is crucial. It's important to note that while time efficiency is an important factor, the accuracy and reliability of the forecasting should also be considered when evaluating the overall performance of these architectures.

As we can see in Figure 2, the custom ANN architecture clearly surpasses the overall performance, especially in the number of successful prediction signals throughout the entire experiment. Additionally, the time required for it to deliver a prediction using the same computational resources is 3.65 to 9.5 times less than that of the compared LSTM architectures.

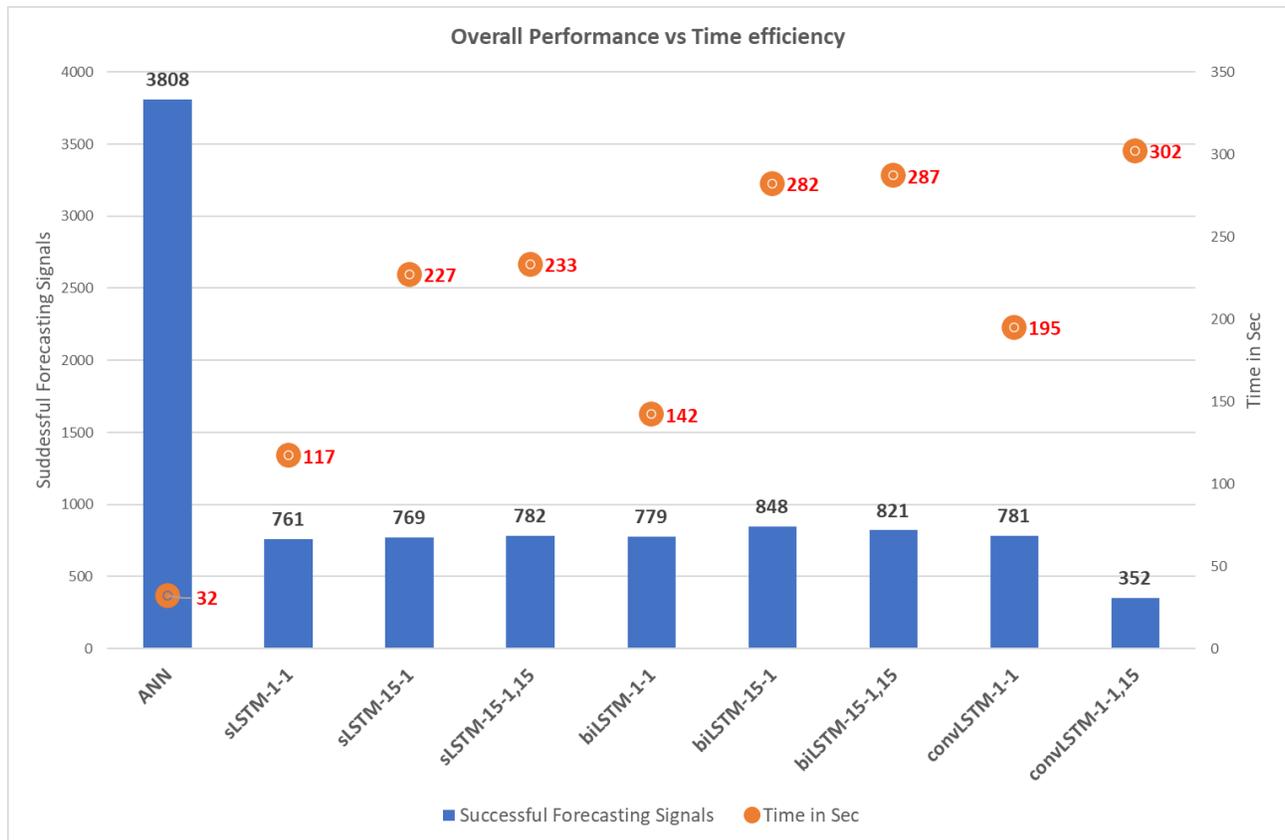

**Figure 2:** Comparative Analysis of Overall Performance Versus Time Expended for Each Architecture

## 5 Conclusions and Further Work

This study conducted a comprehensive comparative analysis of the execution time of various LSTM neural networks and of an ANN specialized architecture for forex market prediction. The results indicate that the ANN specialized architecture not only achieves better results in forex market prediction but also executes using fewer resources and in a shorter time frame compared to LSTM architectures. This finding is significant as it suggests that specialized architectures can offer a more efficient alternative to conventional generic, off the-self, LSTM models for time series prediction in the forex market.

The ANN specialized architecture demonstrated a clear advantage in terms of execution time compared to the LSTM architectures. This efficiency is particularly relevant in the context of forex market prediction, where timely decisions are crucial and improved implementation speed could allow more players to enter the market.

Furthermore, and quite as importantly, the specialized architecture produced a higher number of successful forecasting signals with greater accuracy, indicating its robustness and reliability as a predictive model. The architecture's ability to generate more accurate predictions with fewer resources highlights its substantial potential.

In conclusion, the ANN specialized architecture presents a compelling case for its adoption in forex market prediction tasks. Its superior performance in terms of accuracy, execution time, and resource efficiency positions it as a promising alternative to LSTM neural networks. Future research could delve into the scalability of this specialized ANN architecture, assessing its potential to handle larger datasets and more complex financial forecasting scenarios. Investigating whether the architecture can maintain its efficiency and accuracy when scaled up will be crucial for broader applications. Additionally, exploring its adaptability across different financial markets, beyond forex, could reveal its versatility and potential as a universal tool for financial analysis.

### ACKNOWLEDGMENTS

We acknowledge the comments we have received from anonymous reviewers on earlier versions of this work.